\documentclass{article}
\usepackage[preprint]{corl_2026} 
\usepackage{graphicx}
\graphicspath{{./}}
\usepackage[table]{xcolor}
\usepackage{amsmath,amssymb}
\usepackage{booktabs}
\usepackage{enumitem}
\usepackage{float}
\usepackage{caption}
\usepackage{subcaption}
\usepackage{placeins}
\usepackage{wrapfig}

\definecolor{nicegreen}{rgb}{0.1, 0.6, 0.2}

\newcommand{\OURS}[0]{\textsc{\textsc{Afford2Act}}}

\title{\OURS: How Far Can Sparse Keypoints Go for Robot Manipulation?}

\author{
  Anukriti Singh, Kasra Torshizi, Khuzema Habib, Kelin Yu, Ruohan Gao, Pratap Tokekar \\
  Department of Computer Science \\
  University of Maryland, College Park \\
  College Park, MD, USA \\
  \texttt{\{anukriti, ktorsh, khabib, kyu85, rhgao, tokekar\}@umd.edu}
}

\begin{document}
\maketitle

\begin{abstract}
Vision-based manipulation policies are commonly learned from rich visual observations, typically using dense visual features extracted from full resolution images. It remains unclear how much of this visual information is really needed for effective policy learning. In this paper, we investigate the opposite end of the spectrum: how effective can manipulation policies be if they use as input only a compact visual representation. Sparse 2D keypoints provide a deliberately compact representation, giving the policy spatial information while removing direct access to appearance, texture, and background. We design \textsc{Afford2Act} (\textbf{Affordance-Guided Keypoints to Act}) to study how much manipulation capability can be retained under this representation constraint. An off-the-shelf language-conditioned affordance model identifies the task-relevant region, from which up to 15 object keypoints are selected, together with 4 end-effector keypoints. The policy input is therefore reduced to only 19 2D keypoints (38 coordinates). 
A lightweight gating further decides which keypoints to pay attention to during execution. Across six real-robot manipulation tasks, this compact state outperforms the evaluated RGB, RGB-D, and dense-affordance behavior cloning baselines, degrades less as the number of demonstrations is reduced, and achieves 82\% average success in held-out settings. Replacing learned gating with uniform weights reduces success from 92\% to 52\%. At the same time, the representation depends on reliable affordance localization and keypoint tracking, and remains vulnerable to incorrect masks, occlusion, tracking drift, and distractors.
\end{abstract}


\begin{figure}[!h]
\centering
\includegraphics[width=\linewidth]{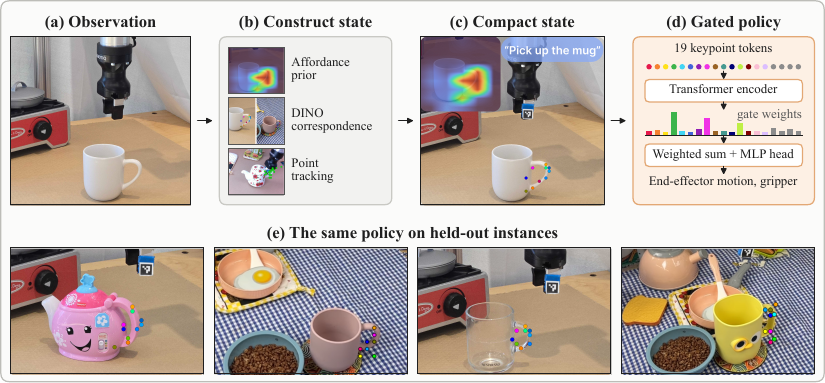}
\caption{\textbf{Overview.} (a) An RGB observation is reduced, using (b) off-the-shelf affordance, correspondence, and tracking models, to (c) a compact state of 19 keypoints: up to 15 on the object's actionable region and 4 on the robot end-effector. (d) The policy encodes the keypoints with a transformer, weights each keypoint with a learned gate (weights shown schematically), and predicts end-effector motion and the gripper command. (e) The same policy on held-out instances.}
\label{fig_1}
\end{figure}

\section{Introduction}

Robot manipulation policies are commonly learned from rich visual observations such as images~\cite{chi2024diffusionpolicy}, point clouds~\cite{dp3}, or semantic features~\cite{uad, wang2024gendp3dsemanticfields}. These representations preserve detailed scene information, but they may also burden a policy with information that is not relevant to the task~\cite{tomar2021learningrepresentationspixelbasedcontrol}. This raises a simple question: \emph{how much of the visual observation does a manipulation policy actually need?}

2D keypoints~\cite{doersch2023tapir, manuelli2019kpamkeypointaffordancescategorylevel} provide a deliberately compact representation. A small set of points gives the policy the spatial structure of the objects being manipulated, while removing its direct access to appearance, texture, and background. This compactness can reduce nuisance visual variation and enables a lightweight policy with lower computational demand, but it also risks removing information that the policy needs. This trade-off is the focus of our work: \emph{can a very small set of task-relevant 2D keypoints retain enough information for effective manipulation and transfer?}

Which points are kept matters as much as how few there are. For a sparse state to support manipulation, the extracted points must lie on regions relevant to the instructed interaction, and each point must keep a consistent identity across demonstrations and object instances. Existing approaches select keypoints from object motion~\cite{track2act, im2flow2act, genflowrl}, learn them jointly with the policy~\cite{zhang2025atkautomatictaskdrivenkeypoint}, or group semantic object keypoints~\cite{wang2025skil, fang2024keypointabstractionusinglarge}. These points describe where an object is or how it moves, but they are not necessarily tied to where the robot should act (Sec.~\ref{sec:related}). In manipulation, relevance is task-conditioned: the same tool supports multiple interactions, and different instructions should highlight different regions (e.g., a brush can be grasped to lift/hold, or oriented and contacted to sweep).

We use \textsc{Afford2Act} (\textbf{Affordance-Guided Keypoints to Act}) to construct and study such a compact state (Fig.~\ref{fig_1}). An off-the-shelf language-conditioned affordance model serves as a semantic prior that restricts keypoint selection to the task-relevant region. Within this region, DINO features~\cite{oquab2024dinov} are used to select a small set of keypoints and to match them across object instances, so that each keypoint index refers to the same object part. The keypoints are tracked over time~\cite{karaev2024cotracker}, and a lightweight transformer policy with a gating layer maps them to robot actions. The resulting state reduces a $224\times224$ image to 19 keypoints (38 dimensions): up to 15 on the object's actionable region and 4 on the robot end-effector (Fig.~\ref{fig_1}). The full image is still used to construct this state, by the affordance and DINO models; our question is whether the policy itself needs it.

We study this compact state on six real-robot manipulation tasks, asking four questions in turn: whether a few semantically rich, automatically extracted keypoints are enough for a manipulation policy, which keypoints should be extracted, how the policy uses them during execution, and when the compact representation fails. The compact state outperforms the evaluated dense-input baselines and degrades less as demonstrations are reduced. Which points are extracted matters beyond compactness alone, and the policy concentrates its weight on a subset of them. The main failures arise when the affordance mask or keypoint tracking is unreliable.

\section{Related Work}
\label{sec:related}
\textbf{Visual Representations for Policy Learning}: Classic approaches rely on raw sensory inputs such as images, depth maps, or point clouds to provide a complete view of the environment~\cite{dp3}. These representations are rich, but they give policies high-dimensional inputs that typically require large amounts of data to learn from. Manipulation-centric and object-centric representations~\cite{uad, jang2017end, florence2018dense, shridhar2024perceiver, sketch2skill, partpolicy, spot} filter out much of the visual clutter and focus policy learning on the features that matter most for action, often relying on detectors or dense inputs from images or 3D reconstructions. We study the far end of this spectrum, where the policy observes only a small set of 2D keypoints.

\textbf{Keypoints and Automatic Anchor Discovery}: Keypoints, when properly chosen, offer a natural low-dimensional representation of an object's essential affordances~\cite{levy2024p3poprescriptivepointpriors, im2flow2act, genflowrl, wang2025skil}. Earlier work often assumed keypoints could be specified by experts or required laborious labeling to maintain semantic consistency across objects~\cite{levy2024p3poprescriptivepointpriors, manuelli2019kpamkeypointaffordancescategorylevel}. More recent methods discover keypoints automatically. Motion-driven approaches identify pixels or point tracks that move significantly when an object is manipulated and learn control from the resulting tracks~\cite{track2act, im2flow2act, genflowrl}; they are effective when object motion is clean, but background motion, such as a person working in the same space, can introduce spurious tracks. Other methods learn which points are most predictive for the task~\cite{zhang2025atkautomatictaskdrivenkeypoint, wen2024anypointtrajectorymodelingpolicy}; ATK, for instance, selects keypoints jointly with the policy, which yields compact representations but can couple the selected points to the training data. Object-centric methods group semantic keypoints for consistent localization across frames and instances~\cite{wang2025skil, fang2024keypointabstractionusinglarge}, but they capture object location and appearance rather than the region relevant to the instructed interaction. VLM-driven pipelines such as MOKA~\cite{liu2024moka} enable zero-shot keypoint discovery through iterative visual prompting. We instead select keypoints on the region relevant to the instructed interaction, and focus on what such a compact state provides for control.

\textbf{Affordance Reasoning in Manipulation}: The concept of affordances has become foundational in robotics, bridging perception and action by focusing on what an object ``invites'' a robot to do~\cite{gibson1979ecological, ciocarlie2007dimensionality}. Early works constructed dense affordance maps~\cite{myers2015affordance, do2018affordancenet, humanaffordance} or trained region classifiers, but these methods faced challenges in scaling and generalization. Vision-language and foundation models~\cite{liu2024moka, li2024ooal} brought a new wave of affordance reasoning, inferring actionable regions from text and image cues. We use such a model only as a semantic prior for selecting keypoints; the affordance map itself is not passed to the policy.


\section{\textsc{Afford2Act}}
\label{sec:method}
\label{sec:problem}
\noindent\textbf{Overview.}
We are given single-view demonstrations $\mathcal{D}=\{(I^{(d)}_{1:T_d},\,\mathbf{u}^{(d)}_{1:T_d},\,A^{(d)})\}_{d=1}^{D}$, where $I^{(d)}_t\in\mathbb{R}^{H\times W\times 3}$ are images, $\mathbf{u}^{(d)}_t\in\mathbb{R}^{m}$ are actions (6-DoF end-effector deltas and a gripper command), $T_d$ is the demonstration length, and $A^{(d)}$ is an affordance text prompt. \textsc{Afford2Act} reduces each image to a small set of task-relevant keypoints (Sec.~\ref{sec:representation}) and learns a policy that acts only on their trajectories (Sec.~\ref{sec:policy}); Fig.~\ref{fig_1} shows the full pipeline.

\noindent\textbf{Notation.}
Each demonstration $d$ is summarized by trajectories of $K$ 2D keypoints,
$\mathbf{P}^{(d)}=\{(x^{(d)}_{t,i},\,y^{(d)}_{t,i})\}_{t=1:T_d,\; i=1:K}$,
where each index $i$ refers to a part within the affordance region (e.g., ``handle tip,'' ``blade edge'').

\subsection{Affordance-Guided Keypoint Representation}
\label{sec:representation}
The keypoints are extracted in four stages, without manual annotation.

\textbf{Affordance Localization:}  
Given the first frame $I_1$ of a demonstration and an affordance prompt $A$, we use an off-the-shelf, pre-trained One-Shot Open Affordance model~\cite{li2024ooal} $f_{\text{aff}}$ to produce a heatmap: $H = f_{\text{aff}}(I_1, A) \in [0,1]^{H \times W}$. This heatmap indicates regions of the scene that correspond to the intended action. To focus only on the actionable surface, we apply a robust quantile threshold (along with mild smoothing) to obtain a binary mask: $M(x, y) = \mathbb{1}\big[ H(x, y) \geq \tau_q(H) \big]$, where $\tau_q(H)$ is the threshold value. This mask restricts where keypoints are selected and matched in the following steps.

\textbf{Reference Keypoint Extraction:}  
On a reference frame (the first frame of the first demonstration), $I_1^r$, we compute dense feature maps using a DINO backbone ($\phi$)~\cite{oquab2024dinov}: $F_r = \phi(I^r_1) \in \mathbb{R}^{H \times W \times D}$. DINO features provide a part-sensitive similarity measure for matching keypoints across instances. We then restrict attention to the ROI~\footnote{ROI stands for Region of Interest, i.e., the cropped region of the affordance mask.} defined by the affordance mask $M$ and cluster the features within that region to obtain $K$ reference anchor points in a reference anchor set $\mathcal{A}_r$:
\[
\mathcal{A}_r = \left\{ (p_i^r,\, \mathbf{f}_i^r) \right\}_{i=1}^{K}, \qquad p_i^r \in \Omega(M),\ \mathbf{f}_i^r = F_r(p_i^r)
\]
where $\Omega(M)$ is the set of pixel coordinates covered by the mask $M$. Unlike manual annotation-based approaches~\cite{levy2024p3poprescriptivepointpriors,liu2024moka}, this step requires no keypoint labels.

\textbf{Cross-instance Correspondence:}  
For the keypoints to be useful across object instances, each index must refer to the same part on every object. Given a new demonstration, we compute a dense feature map from its first frame, $F_t = \phi(I^t_1)$. For each reference anchor keypoint $(p_i^r, \mathbf{f}_i^r) \in \mathcal{A}_r$, we match it in the new frame by evaluating cosine similarity between the reference descriptor and the per-pixel features, restricted to the affordance ROI:
\[
S_i(x, y) = \left\langle \frac{\mathbf{f}_i^r}{\|\mathbf{f}_i^r\|},\, \frac{F_t(x, y)}{\|F_t(x, y)\|} \right\rangle, \quad (x, y) \in \Omega(M_t).
\]
We then select the best-matching location and its confidence as:
\[
p^t_i = \arg\max_{(x, y) \in \Omega(M_t)} S_i(x, y), \quad
c^t_i = \max_{(x, y) \in \Omega(M_t)} S_i(x, y).
\]
This matching ensures that the index $i$ refers to the same functional part (e.g., ``handle tip'') across demonstrations, even as object geometry and appearance vary. By constraining matches to the affordance mask, we reduce spurious matches to visually similar but task-irrelevant regions.


\textbf{Temporal Keypoint Tracking:}  
After initializing all $K$ keypoints in the first frame, we track them over the demonstration using CoTracker~\cite{karaev2024cotracker}. This produces a temporally ordered set of keypoint locations,
\[
\{(p_{1,1},\ldots,p_{1,K}),\; (p_{2,1},\ldots,p_{2,K}),\; \ldots,\; (p_{T_d,1},\ldots,p_{T_d,K})\},
\]
where $p_{t,i}$ denotes the 2D location of keypoint $i$ at time $t$ and $T_d$ is the demonstration length. Joint tracking improves robustness to occlusion (e.g., when the gripper blocks the view of the handle during grasping) by exploiting correlations among keypoints, and preserves index consistency over time. These keypoint trajectories are the only input to the policy (Sec.~\ref{sec:policy}).

\begin{figure}[t]
\centering
\includegraphics[width=\linewidth]{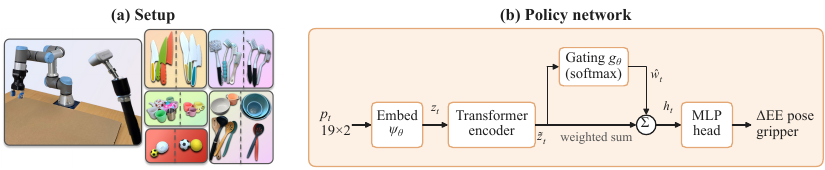}
\caption{\textbf{Experimental setup and policy network.} (a) UR3e arm with a side-mounted RealSense D435i camera, and objects used in the held-out (left column) and seen (right column) scenarios. (b) Each tracked keypoint is embedded by $\psi_\theta$ and contextualized by a transformer encoder $\mathcal{T}_\theta$; a gating network $g_\theta$ produces softmax weights $\hat{w}_t$ that aggregate the tokens into $h_t$, from which an MLP head predicts the end-effector delta action and the gripper command.}
\label{fig:setup_fig_v2}
\end{figure}

\subsection{Policy Design}
\label{sec:policy}

At each inference step $t$, we are given the $K$ tracked keypoints $\{p_{t,i}\}_{i=1}^{K}$. The policy must transform this unordered set into an action while prioritizing the keypoints that are most informative for the instructed manipulation. The network is shown in Fig.~\ref{fig:setup_fig_v2}b. We first embed each keypoint into a token using a learnable embedding function $\psi_\theta$, forming a token set
\[
Z_t = [z_{t,1}, \ldots, z_{t,K}]^\top, \qquad z_{t,i} = \psi_\theta(p_{t,i}).
\]
We then compute contextualized token features with a transformer encoder: $\tilde{Z}_t = \mathcal{T}_\theta(Z_t)$,
so that each keypoint is interpreted in the context of the others (e.g., a point on a knife blade should be treated differently from a point on the handle)~\cite{haldar2024bakuefficienttransformermultitask,zhao2023learningfinegrainedbimanualmanipulation}.

\textbf{Gating network.}
The extracted keypoints need not be equally relevant at every timestep, and some may be redundant or noisy. The policy therefore includes a gating network $g_\theta$ that assigns a non-negative weight to each token, allowing it to down-weight less informative keypoints.
Concretely, we compute per-keypoint scores and normalize them with a softmax:
\[
s_{t,i} = w_g^\top \tilde{z}_{t,i} + b_g, \qquad 
\hat{w}_{t} = \operatorname{softmax}([s_{t,1},\ldots,s_{t,K}]),
\]
where $\hat{w}_t \in \mathbb{R}^{K}$ and $\hat{w}_{t,i} \ge 0$ with $\sum_{i=1}^{K}\hat{w}_{t,i}=1$. We form gated token features and aggregate them into a compact scene embedding: $h_t = \sum_{i=1}^{K} \hat{w}_{t,i}\,\tilde{z}_{t,i}$. The resulting $h_t$ is then passed through a shared MLP action head, which branches into two linear outputs: one predicts the end-effector delta action and the other predicts the gripper command.

\section{Experiments}
\label{sec:experiments}

We study the compact representation on six real-robot manipulation tasks, following the four questions raised in the introduction.

\subsection{Experimental Setup}
\label{sec:setup}

We use a UR3e robot arm and a side-mounted Intel RealSense D435i camera (Fig.~\ref{fig:setup_fig_v2}a). Each task is specified by a natural-language affordance prompt. The affordances include \textit{hold}, \textit{cut\_with}, \textit{sweep}, \textit{pour}, \textit{stir}, and \textit{kick}, and are instantiated into the following tasks: grasp-and-lift, cutting, brushing/sweeping, pouring, stirring, and ball-kicking.

For each task, we train an independent policy from 40 teleoperated demonstrations with randomized object poses, and evaluate it on real-world rollouts under seen and held-out conditions. The held-out conditions include novel instances from the same category, novel categories that share functional parts with the training objects, background and lighting changes, and distractor objects or motion. The objects vary in size, geometry, material, and appearance.

\subsection{Training and Evaluation Protocol}
\label{sec:eval_protocol}

All methods are trained with the same behavior cloning objective and evaluated using the same action space. Keypoint-based methods use a state of $K=19$ keypoints (up to 15 object keypoints and 4 end-effector keypoints, with missing object keypoints masked), so that differences among them reflect the keypoints themselves rather than controller capacity. Unless otherwise stated, each policy is evaluated on 20 real-world episodes per task, and success is reported separately for seen and held-out settings; task-level details are given in Appendix~\ref{app:task_table}. For the semantic subgoal analysis (Table~\ref{tab:semantic_subgoal}), we manually annotate intermediate subgoals on seen-object rollouts for each task, excluding ball-kicking because it does not include a grasping subgoal. For data-scaling experiments, each method is retrained on reduced demonstration subsets.

The pretrained affordance and correspondence modules run only on the first frame of each demonstration. Training one \textsc{Afford2Act} policy from 40 demonstrations takes approximately 15 minutes on a single RTX 2080 GPU.

\subsection{Q1: Are a Few Semantically Rich Keypoints Enough for Manipulation?}
\label{sec:E1_sparse_dense}

We first compare the compact state with policies that observe dense visual inputs. Each baseline is trained on the same demonstrations and action labels:

\begin{itemize}[leftmargin=*, itemsep=1pt]
\item \textbf{ResNet RGB-BC:} A behavior cloning policy trained on raw RGB images using a ResNet-18 visual encoder followed by an MLP action head.

\item \textbf{ResNet RGB-D-BC:} The same architecture as RGB-BC, but trained on 4-channel RGB-D observations using RGB and depth from the RealSense camera.

\item \textbf{UAD-2D-BC:} The same behavior cloning framework, but with RGB inputs replaced by dense affordance masks generated by UAD~\cite{uad}.

\end{itemize}

\begin{figure}[t]
\centering
\begin{subfigure}[t]{0.48\linewidth}
    \centering
    \includegraphics[width=\linewidth]{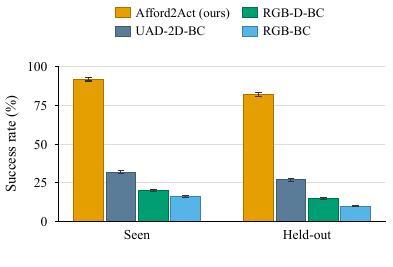}
    \caption{Input modalities}
    \label{fig:success-rate-comparison-modalities}
\end{subfigure}\hfill
\begin{subfigure}[t]{0.48\linewidth}
    \centering
    \includegraphics[width=\linewidth]{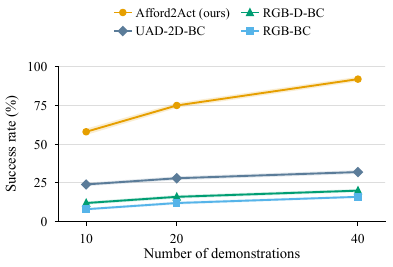}
    \caption{Data scaling}
    \label{fig:data_scaling}
\end{subfigure}
\caption{\textbf{Sparse keypoints versus dense visual inputs.} (a) Total average success rate across input modalities; for each modality we run 20 trials per task and average across tasks, reporting seen and held-out performance separately. (b) Success as the number of demonstrations is reduced; standard deviation bars show variability across tasks.}
\end{figure}

The keypoint policy achieves higher success than the RGB, RGB-D, and dense affordance baselines, with the largest gap in held-out settings (Fig.~\ref{fig:success-rate-comparison-modalities}). Although most of the image is removed, the extracted keypoints carry enough information for these tasks. Dense RGB and RGB-D policies must learn visual invariance and control together from a limited number of demonstrations, which makes them sensitive to background, object appearance, and scene layout. Dense affordance maps are more focused on the task, but still leave the policy to infer the relevant geometry from a full spatial map.

The difference becomes clearer as demonstrations are reduced (Fig.~\ref{fig:data_scaling}). The keypoint policy remains effective with fewer demonstrations, whereas the dense baselines degrade more sharply. This is consistent with the compact state reducing the amount of visual variation the policy must learn from demonstrations.

\begin{table}[t]
\centering
\caption{\textbf{Semantic subgoal analysis on seen instances.} Percentage of annotated rollouts that reach each intermediate stage of the task, rather than only final task success.}
\label{tab:semantic_subgoal}
\small
\begin{tabular}{@{}lcccc@{}}
\toprule
Subgoal (\%) & \textsc{Afford2Act} & UAD-2D & RGB & RGB-D \\
\midrule
Aligned with object & 94 & 80 & 52 & 78 \\
Grasped object & 88 & 75 & 24 & 42 \\
Completed final motion & 86 & 65 & 32 & 28 \\
Robot collided & 2 & 70 & 20 & 12 \\
\bottomrule
\end{tabular}
\end{table}

Task success alone does not show where a policy fails. A semantic subgoal rubric records whether each rollout aligns with the object, grasps it, completes the final motion, or collides with the scene (Table~\ref{tab:semantic_subgoal}). RGB and RGB-D policies can attempt the final task motion before completing a necessary step, such as pouring or cutting before grasping, whereas the keypoint policy more reliably completes early-stage subgoals; its remaining failures are often final contact errors or object slippage.

\subsection{Q2: Which Keypoints Should Be Extracted?}
\label{sec:E2_keypoint_construction}

\begin{figure}[t]
\centering
\begin{minipage}[t]{0.48\linewidth}
\centering
\includegraphics[width=\linewidth]{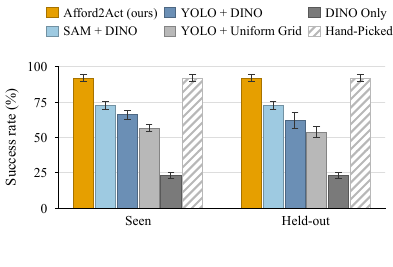}
\caption{\textbf{Keypoint selection.} Total average success rate across keypoint selection methods. All methods use the same number of keypoints and the same policy architecture.}
\label{fig:success-rate-comparison-keypoints}
\end{minipage}\hfill
\begin{minipage}[t]{0.48\linewidth}
\centering
\includegraphics[width=\linewidth]{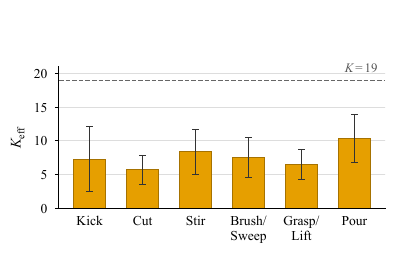}
\caption{\textbf{Effective number of keypoints.} $K_{\text{eff}}$ per task, computed over offline inference on all 40 training demonstrations per task. Although the policy input contains $K=19$ keypoints, the gating concentrates on a smaller effective subset.}
\label{fig:k-eff}
\end{minipage}
\end{figure}

Compactness alone does not determine performance; what matters is which points are kept. We hold the policy and the number of keypoints fixed and vary only how the keypoints are selected:

\begin{itemize}[leftmargin=*, itemsep=1pt]
\item \textbf{DINO Only:} DINO correspondences are computed over the entire image without object or affordance filtering. Tests correspondence alone.

\item \textbf{YOLO + Uniform Grid:} A YOLO detector localizes the target object, and keypoints are sampled uniformly inside the detected bounding box. Tests object localization alone.

\item \textbf{YOLO + DINO:} DINO correspondences are restricted to the YOLO-detected bounding box. Tests object-level correspondence.

\item \textbf{SAM + DINO:} Segment Anything (SAM) produces object masks, and DINO correspondences are sampled within the target-object mask identified using YOLO~\cite{wang2025skil}. Tests object masks.

\item \textbf{Hand-Picked (P3PO-style):} Expert keypoints are manually specified per task following P3PO~\cite{levy2024p3poprescriptivepointpriors}, then tracked using DIFT + CoTracker. An approximate upper bound.

\end{itemize}

Affordance-guided keypoints outperform generic DINO correspondences and the object-centric strategies (Fig.~\ref{fig:success-rate-comparison-keypoints}). Localizing the object is not enough; the points need to lie on the regions relevant to the instructed interaction. Without an interaction prior, DINO-only matching can place points on the background or on task-irrelevant parts of the object. YOLO- and SAM-based selection keeps points on the object, but a bounding box or full mask covers the entire mug, brush, or tool, while the policy needs the handle, edge, rim, or contact surface relevant to the task. Combining the affordance mask with DINO correspondence keeps the points on the task-relevant region and consistent across instances, which matches the hand-picked baseline without task-specific annotation.

\subsection{Q3: Does the Policy Use All Extracted Keypoints Equally?}
\label{sec:E3_effective_keypoints}

The 19 keypoints need not be equally relevant at every timestep: points that matter during approach may differ from those that matter during contact or task completion. To measure how concentrated the gating weights $\hat{w}_{t,i}$ are, we compute the effective number of keypoints:
\begin{equation}
K_{\text{eff}} = \frac{1}{\sum_{i=1}^{K} \hat{w}_{t,i}^2}.
\label{eq:keff}
\end{equation}
A smaller $K_{\text{eff}}$ indicates that the policy concentrates its weight on fewer keypoints.

\begin{figure}[t]
\centering
\includegraphics[width=\linewidth]{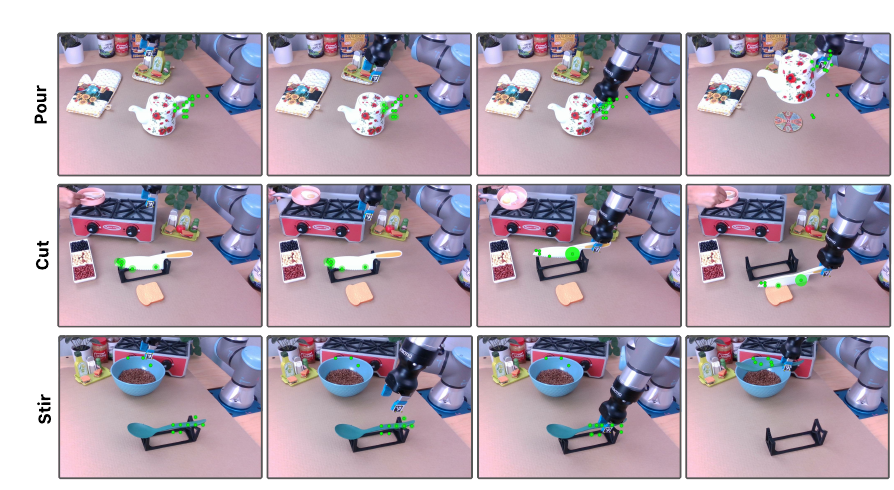}
\caption{\textbf{Rollouts on held-out instances.} Rows show \emph{Pour} (top), \emph{Cut} (middle), and \emph{Stir} (bottom), with each sequence progressing left-to-right from approach $\rightarrow$ interaction $\rightarrow$ completion. Green dots denote affordance-selected keypoints, and translucent ring size indicates gating weight over time. Keypoints remain anchored to task-relevant parts, while the gating weight shifts with phase (e.g., cut: blade tip $\rightarrow$ handle; pour: edge/spout $\rightarrow$ main handle; stir: handle $\rightarrow$ bowl).}
    \label{fig:qualitative_one}
\end{figure}

$K_{\text{eff}}$ is substantially smaller than 19 across tasks (Fig.~\ref{fig:k-eff}), so the policy concentrates its weight on a subset of the extracted keypoints. Qualitatively, the high-weight keypoints shift as the robot moves through approach, contact, and completion (Fig.~\ref{fig:qualitative_one}).

Replacing the learned gating with uniform weights reduces success from $92\%$ to $52\%$, with especially large degradation in cluttered or visually ambiguous settings. Even within an already compact state, learning how strongly each keypoint should influence the action matters.

\subsection{Q4: When Does the Compact Representation Fail?}
\label{sec:generalization}
\label{sec:failures}

\begin{figure}[t]
\centering
{\setlength{\fboxsep}{0pt}\setlength{\fboxrule}{0.4pt}\setlength{\tabcolsep}{2pt}
\begin{tabular}{@{}ccccc@{}}
 & \parbox[b]{0.225\linewidth}{\centering\footnotesize\textbf{Novel Category}\\[1pt]Cut \,(82\%)} & \parbox[b]{0.225\linewidth}{\centering\footnotesize\textbf{Novel Instance}\\[1pt]Pour \,(84\%)} & \parbox[b]{0.225\linewidth}{\centering\footnotesize\textbf{Background}\\[1pt]Kick \,(84\%)} & \parbox[b]{0.225\linewidth}{\centering\footnotesize\textbf{Distractors}\\[1pt]Brush/Sweep \,(78\%)} \\[3pt]
\rotatebox[origin=c]{90}{\footnotesize\textbf{Train}} & \raisebox{-.5\height}{\fbox{\includegraphics[width=0.225\linewidth]{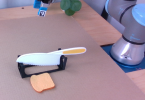}}} & \raisebox{-.5\height}{\fbox{\includegraphics[width=0.225\linewidth]{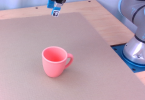}}} & \raisebox{-.5\height}{\fbox{\includegraphics[width=0.225\linewidth]{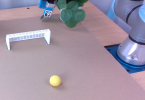}}} & \raisebox{-.5\height}{\fbox{\includegraphics[width=0.225\linewidth]{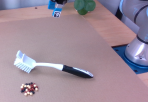}}} \\[2.2pt]
\rotatebox[origin=c]{90}{\footnotesize\textbf{Test}} & \raisebox{-.5\height}{\fbox{\includegraphics[width=0.225\linewidth]{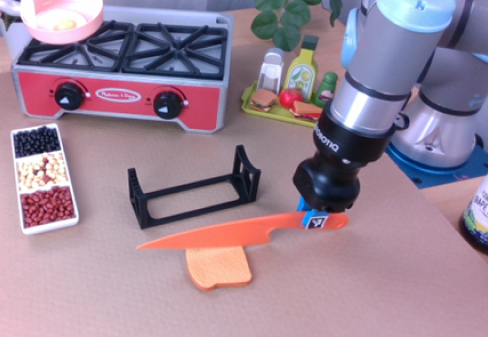}}} & \raisebox{-.5\height}{\fbox{\includegraphics[width=0.225\linewidth]{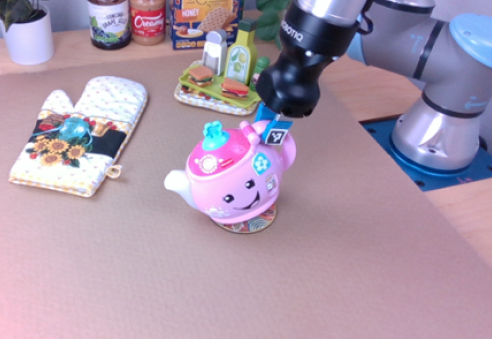}}} & \raisebox{-.5\height}{\fbox{\includegraphics[width=0.225\linewidth]{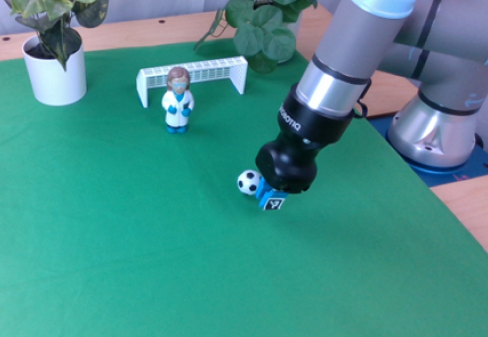}}} & \raisebox{-.5\height}{\fbox{\includegraphics[width=0.225\linewidth]{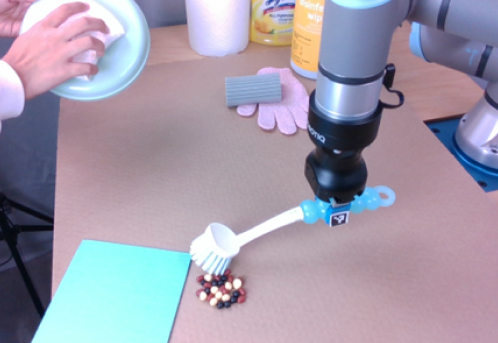}}} \\
\end{tabular}}
\caption{\textbf{Training scenes versus held-out test scenes.} Each column pairs the scene used for training (top) with a held-out test scene (bottom) for one generalization setting, with the task and its success rate under that setting. Test scenes introduce an object category not seen in training (cut with a different blade), a novel object instance (teapot instead of mug), a new background (green tabletop), and distractors (clutter and a person moving in the workspace). The average success rate on seen objects is $92\%$.}
\label{fig:unseen_instances}
\end{figure}

\begin{figure}[t]
\centering
\includegraphics[width=\linewidth]{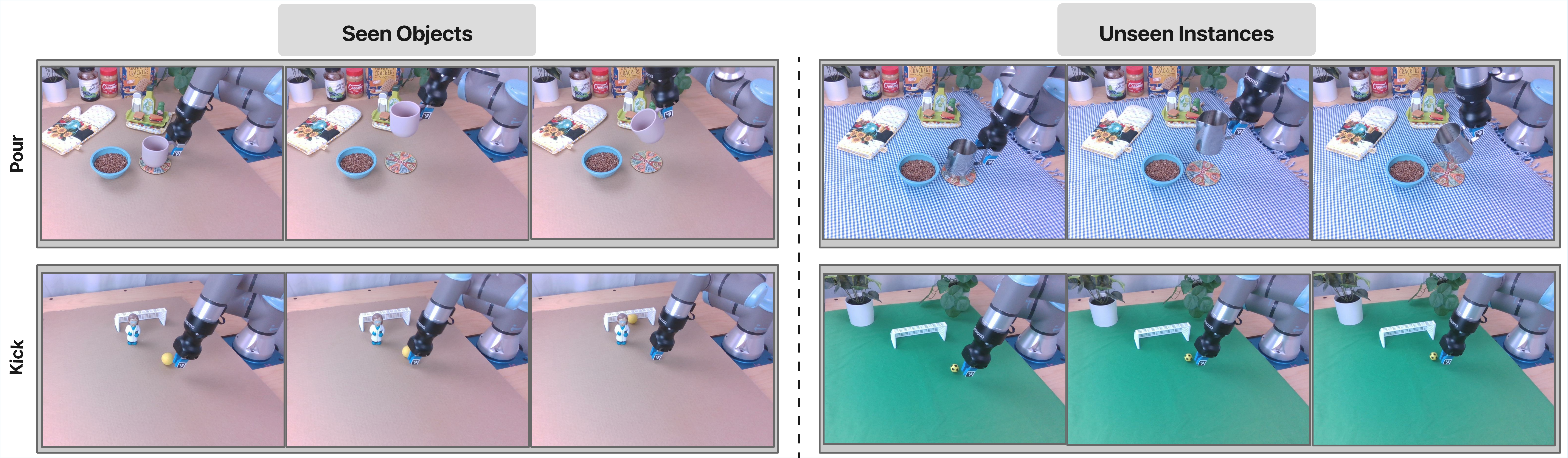}
\caption{\textbf{Qualitative generalization on two tasks.} Rows show rollouts for \emph{Pour} (top) and \emph{Kick} (bottom). \textbf{Left:} seen training objects. \textbf{Right:} held-out settings with new backgrounds and, for \emph{Pour}, novel object instances. Each panel shows three key frames (setup, approach, completion). No test-time adaptation is used.}
    \label{fig:qualitative}
\end{figure}

We evaluate transfer under four held-out settings: novel instances, novel categories that share functional parts with the training objects, background or lighting changes, and distractors (Fig.~\ref{fig:unseen_instances}). Success remains high for novel instances ($84\%$), novel categories ($82\%$), and background changes ($84\%$), compared with $92\%$ on seen objects. This suggests that the policy is not strongly tied to object appearance or scene layout: when a test object preserves a handle-like or contact-relevant region, the keypoints remain aligned with the parts needed for control (Fig.~\ref{fig:qualitative}).

Performance drops most under distractors ($78\%$), and the failures we observe fall into two groups. The first are limitations of the representation itself. The keypoints are 2D and come from a single external RGB view, so they inherit its partial observability: heavy occlusion or viewpoint ambiguity can leave the relevant points hidden or ambiguous, and visual cues that are removed from the policy's input cannot be recovered downstream.

The second are failures of the modules that construct the state. The primary failure mode is an inaccurate or incomplete affordance mask, which can omit the true contact area or include distracting regions, leading to unstable keypoints and incorrect control; Appendix Fig.~\ref{fig:failure_mask} shows an example in which the gripper collides with the spatula. Correspondence errors, tracking drift under contact-induced object motion, and distractors that interfere with affordance localization or tracking have similar effects (Appendix Table~\ref{tab:failure_taxonomy}). When the keypoints are reliable, the policy acts well from the compact state; these failures arise upstream, in constructing it. A preliminary two-stage re-anchoring result is reported in Appendix~\ref{app:reanchoring}.

\section{Conclusion}
We studied how much manipulation capability is retained when visual observations are compressed into 19 task-relevant 2D keypoints. Across six real-robot tasks, a policy observing only these keypoints outperformed the evaluated dense-input baselines, degraded less with fewer demonstrations, and transferred to held-out objects, categories, and backgrounds. Which points are extracted mattered as much as how few there are.

\textbf{Limitations.} We study this compact representation from a single external camera view. Because the policy observes only the extracted 2D keypoints, a keypoint that becomes occluded, for example when the gripper covers the handle during grasping or when a person moves through the workspace, leaves the policy without the information it carried (Sec.~\ref{sec:generalization}). A single viewpoint also ties the keypoints to that camera's perspective, so their usefulness depends on where the camera is placed. A natural direction is to transfer 2D keypoints between viewpoints, so that a point hidden in one view can be recovered from another while the state remains equally compact.

\bibliography{bibliography}

\appendix
\section{Additional Experimental Details}
\label{app:task_details}

This appendix provides task-level details, a taxonomy of the failures observed during real-world evaluation, and a preliminary analysis of multi-phase re-anchoring.

\subsection{Task Details}
\label{app:task_table}

Table~\ref{tab:task_details_appendix} lists the tasks, their affordance prompts, and the number of training demonstrations.

\begin{table}[h]
\centering
\caption{\textbf{Tasks.} Each task is trained as an independent policy from 40 teleoperated demonstrations.}
\label{tab:task_details_appendix}
\small
\begin{tabular}{@{}lcc@{}}
\toprule
Task & Affordance prompt & Training demonstrations \\
\midrule
Grasp/Lift & \textit{hold} & 40 \\
Cut & \textit{cut\_with} & 40 \\
Brush/Sweep & \textit{hold}, \textit{sweep} & 40 \\
Pour & \textit{pour} & 40 \\
Stir & \textit{hold}, \textit{stir} & 40 \\
Kick & \textit{kick} & 40 \\
\bottomrule
\end{tabular}
\end{table}

\subsection{Failure Taxonomy}
\label{app:failure_taxonomy}

Table~\ref{tab:failure_taxonomy} summarizes common failure modes observed during real-world evaluation. The most frequent failures arise from perception errors in the initial affordance mask, tracking drift under occlusion or contact, and execution-level failures such as object slippage.

\begin{figure}[h]
\centering
\includegraphics[width=0.45\linewidth,trim=458 30 2 0,clip]{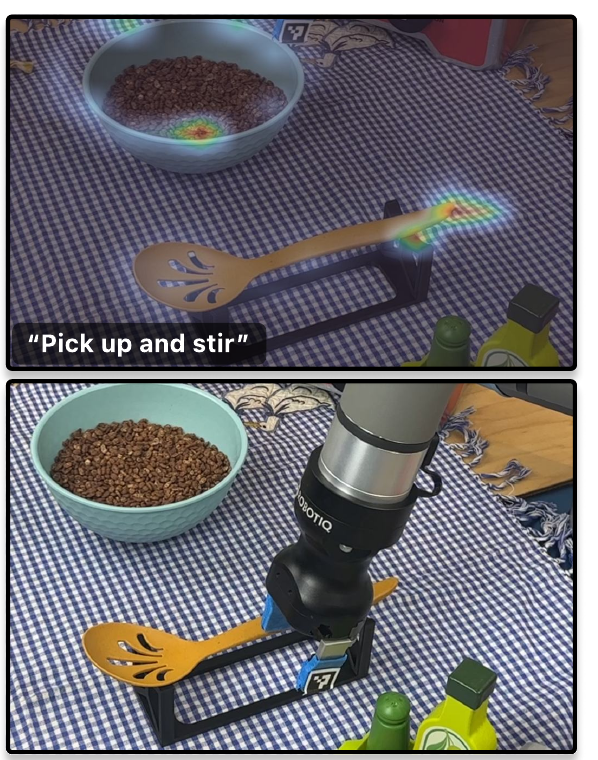}
\caption{\textbf{Failure from weak affordance masking.} On the prompt \textit{``pick up and stir''}, incorrect keypoints lead the gripper to collide with the spatula.}
\label{fig:failure_mask}
\end{figure}

\begin{table}[t]
\centering
\caption{Failure taxonomy observed during real-world evaluation.}
\label{tab:failure_taxonomy}
\resizebox{\textwidth}{!}{
\begin{tabular}{p{4.2cm}p{4.4cm}p{4.4cm}p{4.4cm}}
\toprule
Failure Type & What Happens & Likely Cause & Potential Mitigation \\
\midrule
Affordance mask error & Keypoints are initialized on the wrong region or miss the true contact region. & The affordance model predicts an incomplete or incorrect actionable region. & Confidence-aware keypoint filtering, mask-noise augmentation, or online re-anchoring. \\
Tracking drift & Keypoints move away from the intended object part during execution. & Occlusion, contact-induced motion, or viewpoint ambiguity from the single external camera. & Periodic re-detection, confidence-based tracking checks, or multi-view sensing. \\
Correspondence error & A keypoint index corresponds to an inconsistent functional part across instances. & DINO similarity can be ambiguous when visually similar regions exist. & Stronger ROI filtering, multiple candidate matches, or descriptor confidence thresholds. \\
Execution/contact failure & The robot reaches the relevant region but fails the final interaction. & Grasp slippage, insufficient contact, or small control errors. & Improved gripper/contact control or additional demonstrations near failure cases. \\
Distractor interference & The policy attends to or tracks visually irrelevant objects or motion. & Distractors affect affordance localization or point tracking. & Distractor augmentation, confidence filtering, or re-anchoring during execution. \\
\bottomrule
\end{tabular}
}
\end{table}

\subsection{Two-Stage Re-Anchoring for Multi-Phase Tasks}
\label{app:reanchoring}

We include a preliminary example of two-stage re-anchoring on the \textit{Stir} task. This task contains multiple phases: grasping the tool, moving to the bowl, and executing the stirring motion. After detecting completion of the grasping phase, we update the affordance prompt and re-anchor a subset of keypoints for the second phase. This experiment achieved $90\%$ success on \textit{Stir}, suggesting that online keypoint refresh may be useful for multi-phase manipulation. Since this experiment is currently limited to one task, we treat it as supplementary analysis and leave broader evaluation to future work.


\end{document}